\documentclass[letterpaper, 10 pt, conference]{ieeeconf}  
\usepackage[hidelinks,colorlinks=true,linkcolor=blue,citecolor=blue]{hyperref}
\usepackage[
backend=bibtex
]{biblatex}
\usepackage{graphicx}
\usepackage{mathtools}
\usepackage{subfigure}
\AtBeginBibliography{\fontsize{7pt}{7pt}\selectfont}
\usepackage{caption}
\IEEEoverridecommandlockouts                              

\title{\LARGE \bf
Self-supervised cost of transport estimation for multimodal path planning}

\author{Vincent Gherold$^{1}$, Ioannis Mandralis$^{1*}$, Eric Sihite$^{1}$, Adarsh Salagame$^{2}$, Alireza Ramezani$^{2}$, Morteza Gharib$^{1}$ 
\thanks{Research was conducted at CAST at Caltech}
\thanks{$^{1}$Aerospace Engineering Department, California Institute of Technology, 1200 E California Blvd, Pasadena, CA, USA
        {\tt\small mgharib@caltech.edu}}%
\thanks{$^{2}$Electrical and Computer Engineering Department, Northeastern University, 360 Huntington Ave, Boston, MA,
USA.
        {\tt\small a.ramezani@northeastern.edu}}%
\thanks{$*$ Corresponding author: imandralis@caltech.edu}}

\begin{document}

\maketitle
\thispagestyle{empty}
\pagestyle{empty}

\begin{abstract}

Autonomous robots operating in real environments are often faced with decisions on how best to navigate their surroundings. In this work, we address a particular instance of this problem: how can a robot autonomously decide on the energetically optimal path to follow given a high-level objective and information about the surroundings? To tackle this problem we developed a self-supervised learning method that allows the robot to estimate the cost of transport of its surroundings using only vision inputs. We apply our method to the multi-modal mobility morphobot (M4), a robot that can drive, fly, segway, and crawl through its environment. By deploying our system in the real world, we show that our method accurately assigns different cost of transports to various types of environments e.g. grass vs smooth road. We also highlight the low computational cost of our method, which is deployed on an Nvidia Jetson Orin Nano robotic compute unit. We believe that this work will allow multi-modal robotic platforms to unlock their full potential for navigation and exploration tasks. 


\end{abstract}

\section{INTRODUCTION}
To effectively navigate their environment, robots must be able to identify and understand their surroundings. For instance, a robot attempting to navigate a region composed of dense vegetation or slippery materials such as sand must have some understanding of how difficult each type of terrain is to traverse. Indeed, \textit{traversability}, defined as the difficulty of moving through a specific region based on its geometric and semantic properties \cite{surveytrav}, is a crucial challenge in robotics. Understanding the environment and evaluating ones ability to traverse it is commonly referred to as \textit{traversability estimation}. To push the current frontiers of robotic autonomy, developing efficient traversability estimation algorithms is key. The need for such algorithms is further enhanced due to the current boom of multi-modal robotic platforms. Multi-modal robots are autonomous systems that are able to negotiate a wider range of terrains by switching locomotion modes. For example, a ground-based multi-modal robot such as ANYmal on Wheels \cite{anymal} may be able to traverse sandy terrain by crawling to increase the contact surface, while driving on paved pathways to maximize speed.


Classical methods were the foundation of early research in traversability estimation. Some of these methods use Bayesian probabilities to estimate traversability, such as \cite{probtravmap}, \cite{Pixel}. However, these methods require manual feature engineering and are usually designed for specific terrain classes and may not be easily generalized \cite{surveypoorfeature}.

\begin{figure}[t]
\centering
\includegraphics[width=0.48\textwidth]{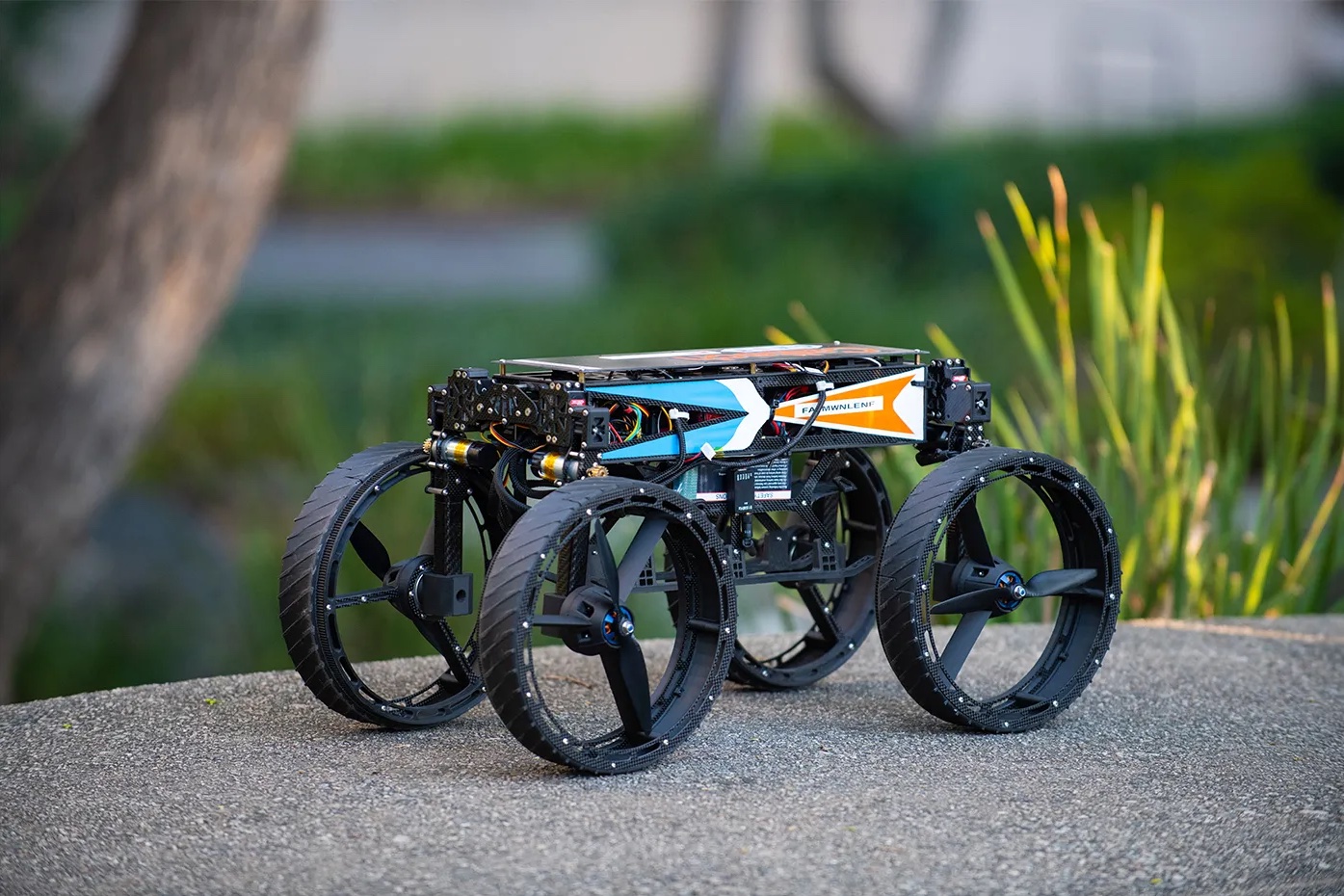}
\caption{Close-up view of our robot platform, the M4 robot, capable of multiple modes of locomotion, including driving, flying, and walking. The robot is equipped with an RGB-D camera along with an embedded companion computer.}
\label{fig:m4}
\end{figure}
Supervised learning based methods have allowed for more accurate traversability estimation. For example \cite{frey2024roadrunner}, \cite{Maturana-2017-102768}, \cite{TNS} use semantic segmentation methods to detect what type of material the robotic platform is driving on. However, these methods have two major drawbacks. Firstly, creating the dataset requires a considerable amount of human-intensive labeling effort. Secondly, the association between classes and terrain can create ambiguity due to the predefined set of classes chosen during the labeling process. For example, if the robot encounters two types of the same terrain, one traversable and the other not, a single class definition for this terrain would make it difficult for the robot to distinguish between them, leading it to incorrect decision making.
\begin{figure*}[ht]
\centering
\includegraphics[width=\textwidth]{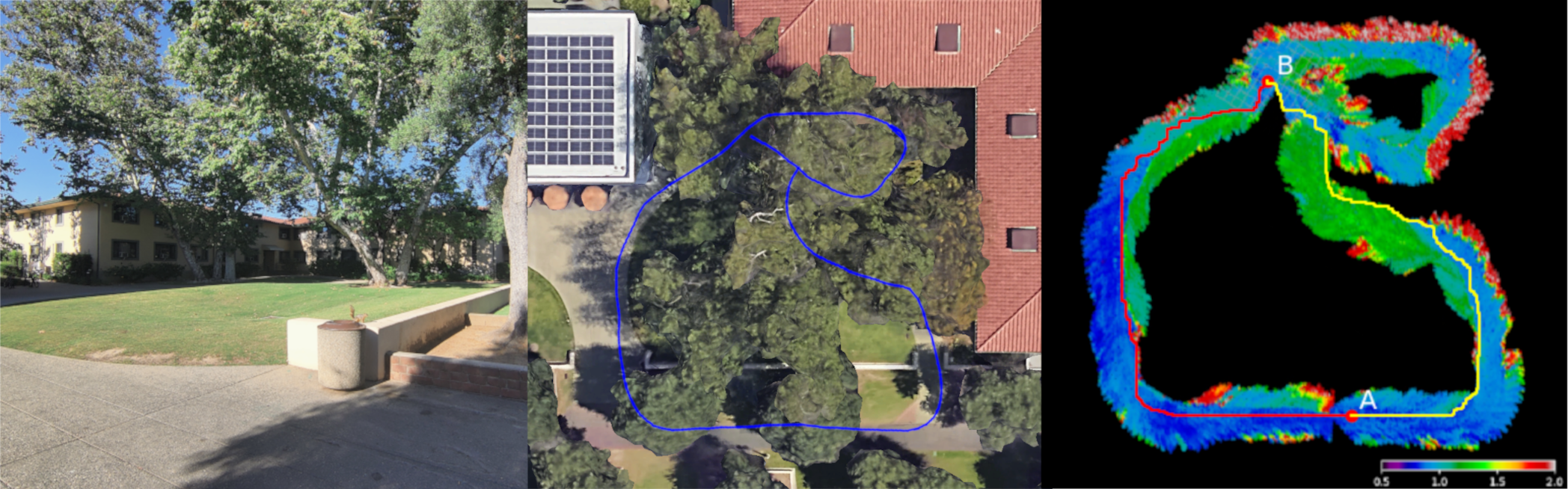}
\caption{Results of our self-supervised COT. The left image shows the terrain that the robot will encounter, composed of road, grass, and vegetation. The middle image displays the trajectory taken by the robot, while the right image presents the resulting COT map, colored using a colormap ranging from 0.5 to 2. On this  COT map, the A* algorithm has been applied to illustrate how COT influences the optimal path to minimize energy consumption along the trajectory. The red path represents the most efficient route, with a total aggregate COT of 817 over a distance of 59 meters. In contrast, the yellow path is a suboptimal solution with a COT of 839 and a distance of 56 meters. Despite being longer, the red path is more energy-efficient because it predominantly follows the road. A video of the mapping is available at \url{https://www.youtube.com/watch?v=tnxdjiAG2Sc}}
\label{fig:results}
\end{figure*}

To address all these challenges self-supervised traversability estimation methods have been developed \cite{frey2023fast}, \cite{seo2024metaverse}, \cite{howfeel}, \cite{selfsupervised}. The principle behind these, involves defining a traversability metric, collecting data while the robot navigates its environment and automatically labeled to create a weakly labeled dataset. The automatic labeling is commonly done after data collection by generating labels using an external sensor (e.g., a LIDAR) or by projecting the robot's trajectory onto the captured images from the robot's camera.
Unlike traditional methods, this approach does not explicitly define any classes; instead, it creates a mapping between features in the data and the traversability metric.

The main contribution of this paper is the development of a novel self-supervised method for traversability estimation using the COT (cost of transport) as a traversability metric, which quantifies the energy efficiency for a robot to move between two points in space \cite{Gabrielli1950}. We demonstrate the benefit of using COT for traversability estimation in the context of multi-modal path planning, allowing to unlock the high degree of locomotion plasticity of M4. As an example, our robot could crawl on rocky terrain to improve stability and drive on smooth road for maximum efficiency. Our method is highly scalable and can be applied to any robotic platform. The framework predicts the cost of transport directly in a BEV (Bird's Eye View) map and integrates the estimation into a single global map. This method has been demonstrated in multiple challenging environments such as grass, rocky terrains and roads.

\begin{figure*}[ht]
\centering
\includegraphics[width=\textwidth]{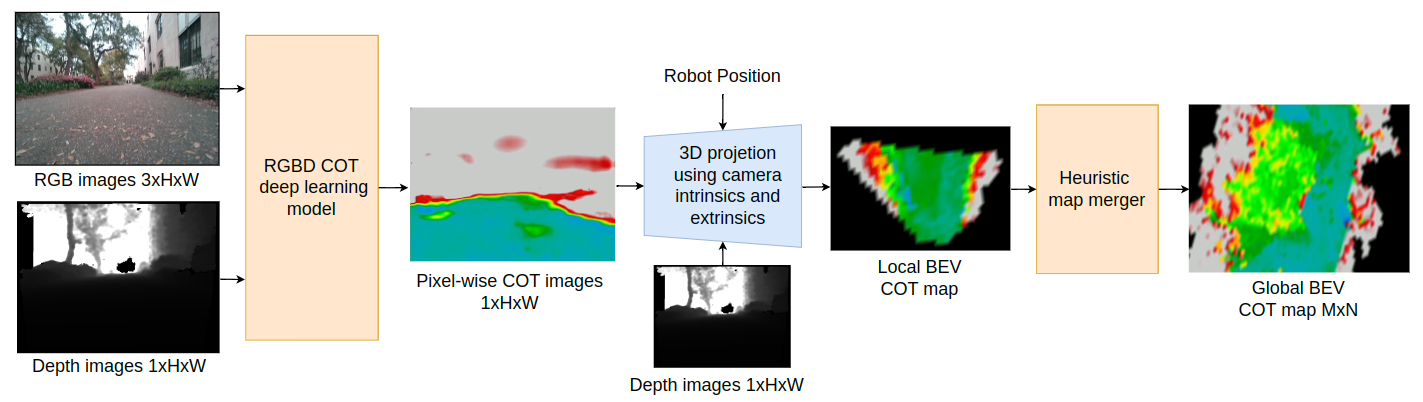}
\caption{Our RGBD COT model takes as inputs an RGB image and depth image and outputs a pixel-wise COT image. RGB and Depth are of size $3\times H\times W$ and  $1\times H\times W$ respectively. Then the pixel-wise COT image is projected in a local BEV map using the robot position and the depth image. A heuristic map merger combines all the local BEV maps into a global map.}
\label{fig:overallpipeline}
\end{figure*}

\section{Related work}
Self-supervised learning is a paradigm in which labels are automatically generated from raw collected data without human intervention. As discussed in \cite{scate}, \cite{frey2023fast}, \cite{selfsupervisiononly}, \cite{wayfast},  \cite{reconstruct}, \cite{gasparino2024wayfaster}, and \cite{seo2024metaverse}, RGBD (RGB + Depth), IMU (Inertial Measurement Unit), and LIDAR  (Light Detection and Ranging) have been identified as suitable candidates for self-supervised methods. These sensors, combined with SLAM (Simultaneous Localization and Mapping), allow the robot to localize itself in space and generate labels automatically.
\cite{scate} demonstrates the effectiveness of self-supervised learning by employing PU (Positive-Unlabeled) learning \cite{PU}, a type of semi-supervised machine learning, to train the deep learning model to distinguish binary classes. 
In contrast, the approach in \cite{reconstruct} addresses the problem through anomaly detection based on reconstruction error. This method utilizes an autoencoder network trained on masked vehicle trajectory regions to identify out-of-distribution scenarios, significantly highlighting non-traversable areas.
Similarly, \cite{selfsupervisiononly} and \cite{jung2024vstrong} generate their labels using self-supervision but enhance performance through contrastive loss - a metric learning objective used
to train models to learn feature embeddings by pulling similar data points closer together and pushing dissimilar points farther apart in the embedding space. According to \cite{constrative},
contrastive loss is highly robust to corruption and stable across different hyperparameter settings.

While the previously mentioned methods leverage geometric and semantic information, \cite{wayfast} and \cite{gasparino2024wayfaster} incorporates temporal information to predict a local traversability BEV (Bird’s Eye View) map around the robot. This map even includes traversability information about terrains that are not visible to the camera due to the temporal deep learning architecture. They define the traversability metric as the traction coefficient from the robot’s kino-dynamic model. This coefficient is directly related to how likely the robot is to slip on the terrain. Additionally, they used a RHE (Receding Horizon Estimator) to label the recorded data. These two works demonstrate their robustness in challenging unstructured outdoor environments.

Online methods that require no prior data have also emerged. For example, \cite{frey2023fast} introduces a self-supervised learning system for traversability that adapts from a short-field demonstration. This system uses features of a visual transformer called DINO \cite{dino} and includes an online supervision pipeline running on the robot. Not all data can be labeled, so a confidence estimation mechanism is introduced to further label the data and identify potential traversable elements. While this method seems attractive, it requires time to adapt to any new environment. Having an efficient offline-based model allows the robot to operate autonomously without needing any teaching examples when deployed.

\section{Methods}
The goal of this work is to create a computer vision pipeline that allows the robot to navigate autonomously in complex, and unstructured environments. Inputs available at inference time are RGBD images and their positions in the 3D world, through SLAM. This combination offers an optimal compromise because RGB images effectively capture the semantic features, while depth data provide geometric information about the environment's 3D structure. The pipeline is composed of a deep learning model that learns the mapping between a traversability metric and features present in the input data. At training time, additional data is required in order to create the training labels. In addition to the previously mentioned input data, speed, current data, voltage data, and occupancy point clouds for our label generation and label augmentation methods described later.

Previous work on our robotic platform (M4 robot \cite{m4}, \cite{Sihite2023}, \cite{Mandralis2023} as shown in Figure \ref{fig:m4}) for optimal path planning has focused on minimizing the energy required to travel between two points in the environment. Following this principle, the COT is selected as the optimal metric for our pipeline. This metric allows us to quantify the efficiency of the robot’s movement across different terrains. The COT is defined as follows:
\begin{equation}
COT = \frac{E}{mgd} = \frac{P}{mgv}
\end{equation}
\noindent where $E$ is the total energy consumed by the robot, $m$ is the mass of the robot, $g$ is the acceleration constant, $d$ is the distance traversed, $P$ is the power consumption, and $v$ is the velocity of the robot. This metric is dimensionless and continuous. 

\subsection{Overview of the pipeline}
The general pipeline is summarized in Figure \ref{fig:overallpipeline} composed of RGBD COT deep learning model, and Heuristic Map Merger. The choice of using a 2D COT image prediction is driven by two primary reasons. First, the goal of this work is to enable real-time deployment on our robotic platform. Predicting 2D COT images simplifies the task for the model, potentially allowing for a smaller and more efficient model. By leveraging 2D COT images, the model can more easily associate specific features in the images with typical traversability metrics. Second, the task of predicting the COT image is analogous to pixel-wise semantic segmentation in computer vision. Techniques and deep learning models developed for semantic segmentation can be applied to achieve high performance.

\subsection{Data collection}
Collecting high-quality data is an essential step for training our RGBD COT deep learning model. The data collection is organized using three main components: RGBD camera, RTAB-Map \cite{RTabmap} SLAM (Simultaneous Localization and
Mapping), and RVIZ \cite{rviz} for visualization and debugging purposes.
During the data collection process, RGBD keyframe of the RTAB-Map SLAM are captured. A keyframe is a selected frame from a sequence of
RGBD images that serves as a reference point for the SLAM and capture significant changes in the environment. By keeping only these keyframes, we maximize the amount of new information added to the dataset while minimizing the required storage space.
Additionally, we saved the reconstructed point cloud from the RTAB-Map algorithm. The position of each keyframe in the point cloud is also stored. Lastly, speed, current and voltage data are also collected.

\subsection{Label generation}
The goal is to generate a pixel-wise continuous value of size $1 \times H\times W $, representing a COT image label. Traversable areas, such as grass and road, will be assigned a COT value, while non-traversable areas will be assigned an arbitrarily high COT value. To generate the labels, our process relies on three key assumptions: 
\begin{itemize}
\item (1) Areas where the robot has traveled can be assumed to be traversable.
\item (2) Elements in a predetermined rectangular region above the robot are considered non-traversable, because the robot is in driving mode. 
\item (3) Similar terrain features correspond to similar traversability. By utilizing the robot’s trajectory data and the terrain features encountered (specifically, parts of the image where the robot has traveled).

\end{itemize}




From assumption (1), generating the labels for the areas the robot traversed during the data collection phase entails first computing the COT along the traversed trajectory using a spatial moving average. This gives us mean COT values which can then be projected back into each RTAB-Map SLAM keyframe image, thus populating the images with projected COT values.  The spatial moving average is computed by converting the RTAB-Map SLAM path into a triangle mesh representation with width corresponding to the robot width. The COT of each triangular mesh element is computed by taking an average of the power consumed by the robot at that position in time over a brief horizon period of 5 meters of trajectory. The COT along the 3d trajectory, as well as the point cloud captured from RTAB-Map is rendered onto a single image using Open3D \cite{Open3D}. The virtual camera used in Open3d for rendering has the same intrinsic and extrinsic parameters as the real RGBD camera.

However, using the above method results in COT labels being extracted from the traversed trajectory only. From assumption (2), we can further improve our label generation procedure by extracting non-traversable labels such as walls, dense vegetation, and rocks from the captured point cloud. To reflect non-traversability, these regions will be labeled a high arbitrary COT. Having these new labels will help the model differentiate traversable and non-traversable features in the data. Finally, parts of the rendered image that are not labeled are assigned an unknown COT. Vectorization and multiprocessing are used to accelerate the rendering and the processing of label generation.

\begin{figure}[h]
\centering
\includegraphics[width=0.485\textwidth]{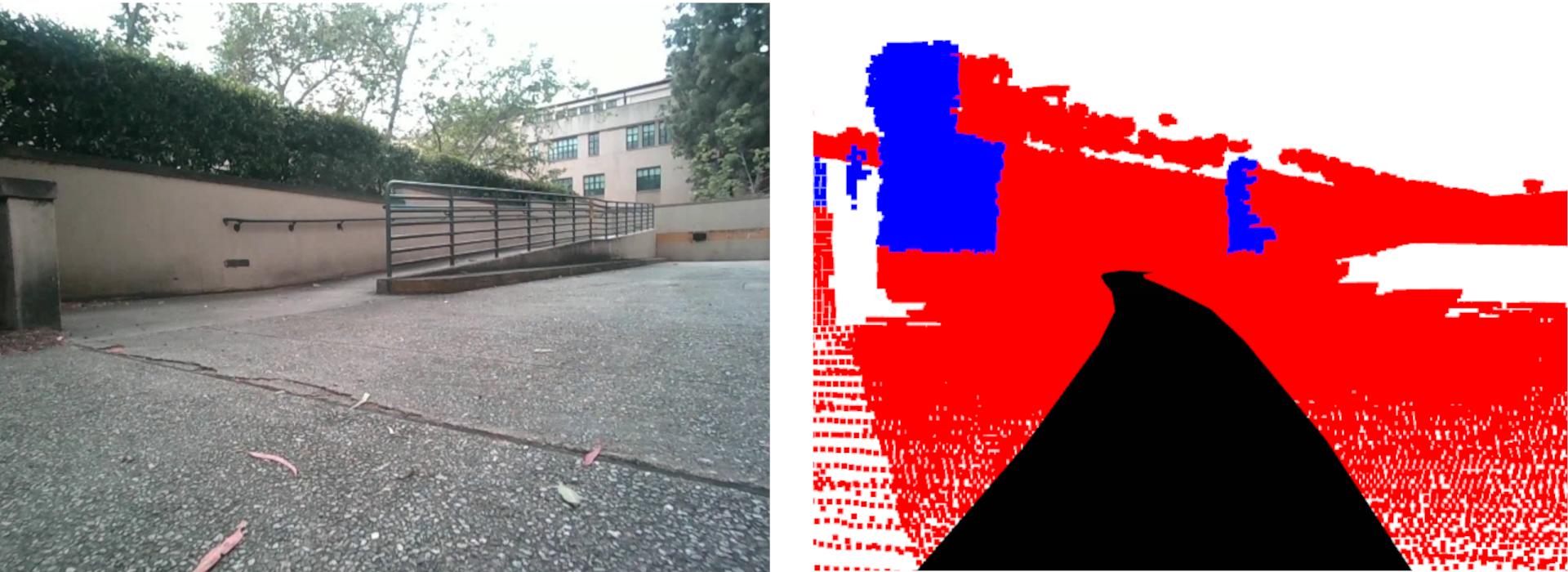}
\caption{The left image displays the camera's view, while the right one shows the rendered world. The elements have been colored for visualization purposes. The triangle mesh COT is represented in black, the point cloud in red, and the non-traversable elements from assumption (3) are colored in blue. The path length and the region size for the blue points have been determined using the maximum depth distance of the camera. On the right image, blue pixels would be assigned a high COT, black a low COT, and the red and white would be assigned an unknown COT.}
\label{fig:open3Dview}
\end{figure}

\subsection{Label Augmentation}

Using the label generation procedure results in only around 34.06\% of the image being labeled with COT values. This is insufficient for training accurate models, as shown in the ablation study (see next Section). Since we have assumed that similar terrain corresponds to similar COT values as stated in assumption (3), we can use a segmentation model to divide each image into regions according to the terrain. To achieve this, we employ a novel segmentation model called SAM (Segment Anything Model) \cite{segmentanything}. This segmentation model is used to identify the local terrain (mask) the robot is currently traversing. If the projected path on the image is included in this mask but not fully, the label of the projected path is extended to the full mask. 

The process is as follows: we use the segmentation model to create masks of similar terrain in the image. We then overlay the projected path onto the segmented masks and  retain the masks which intersect the projected path. The COT of each mask is computed as the average COT of the intersection between the mask and the projected trajectory label or between the mask and the non-traversable label. Thus, each selected mask now has a unified COT and the rest of the image is considered as unknown. This label extension method results in another 18.06\% of the image being assigned labels.


At this point, most of the remaining unknown labels are non-traversable. To identify these non-traversable unknown labels we employ a similar approach to \cite{frey2023fast}. In this approach, the authors mine information from the still unlabeled data by using a small auto-encoder network trained to reconstruct only traversable regions. Thus, the model reconstruct these regions well and non-traversable regions poorly. For example, grass that has been traversed by the robot will be reconstructed well, but a tree that never intersects the projected path will not be trained on, resulting in poor reconstruction. Thus, using this reconstruction metric, we can distinguish the non-traversable parts from the traversable parts in the unlabeled data.



\begin{figure}[h]
\centering
\includegraphics[width=0.48\textwidth]{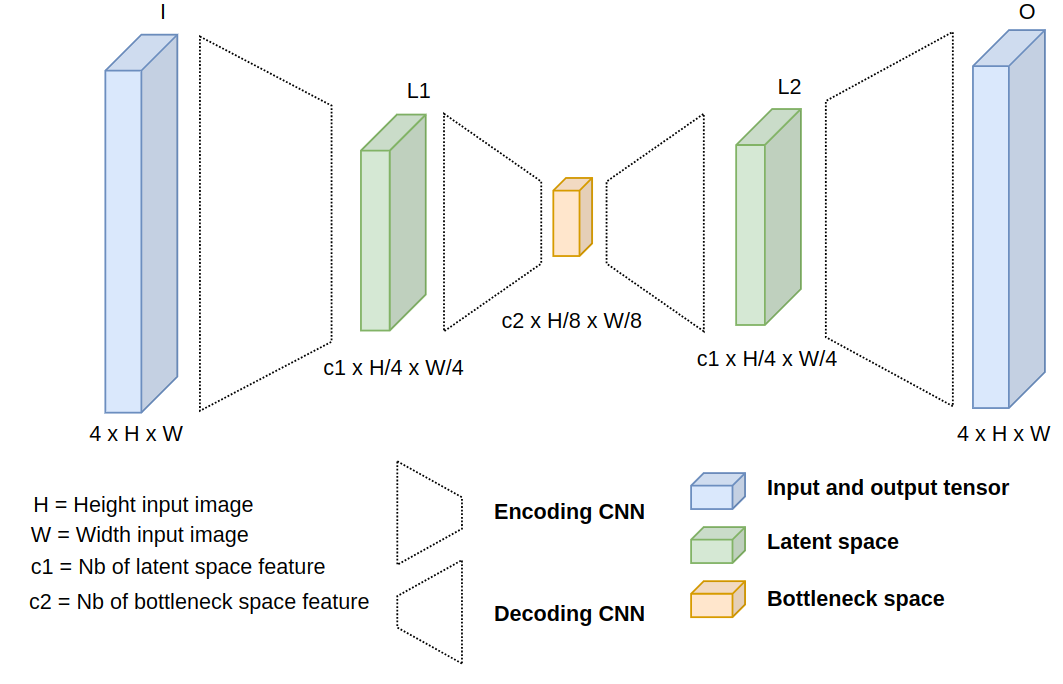}
\caption{Autoencoder architecture for labeling unlabeled data. It takes as input RGBD images in a $4\times H \times W$ tensor. This architecture is composed of two CNN (Convolutional Neural Network) autoencoders, nested one into the other. $I$ denote the input data, $L_1$ the first latent space, $L_2$ the second latent space and $O$ the output space.}
\label{fig:confidencemodel}
\end{figure}

The autoencoder used in \cite{frey2023fast} has been adapted to our task. The inputs and outputs are the same and are composed of the RGBD image concatenated into a $4\times H \times W$ images. As shown in Figure \ref{fig:confidencemodel}, the autoencoder architecture consists of two encoders and two decoders organized one after the other. Therefore, the model shows two latent spaces $L_1$, $L_2$ and a bottleneck space. The main loss is a MSE loss on the inputs and outputs of the model (in blue on figure \ref{fig:confidencemodel}). The model is trained on only the traversable labeled part of our COT image by applying a mask to the main loss. An auxiliary reconstruction MSE loss ensures that the two latent spaces (in green on figure \ref{fig:confidencemodel}) remain close. The total loss is composed of the reconstruction loss and the auxiliary loss. The fact that we first encoded the input data into these two latent spaces helps to compensate for the heterogeneous complexity of the terrain. These latent spaces represent pixel-wise terrain features. For example, a concrete ground of one unified color can be easy to reconstruct but a grass terrain can be harder due to more complex features. 

The mean mask loss is defined as:
\begin{equation}
   \mu_{L_1}(S_i) = \frac{1}{|S_i|} \sum_{(x,y) \in S_i} \sum_{j=1}^{c1} L_1(j,x,y) 
\end{equation}
where $S_i$ represents a mask from SAM in the batch, $|S_i|$ is the number of mask in the batch. $L_1(j,x,y)$ represent the first latent space denoted as $L_1$ of size $c1 \times \frac{H}{4} \times \frac{W}{4}$.

A similar approach can be done for in the second latent space leading to $\mu_{L_2}(S_i)$. Then, the squared error for each mask $ S_i$ is computed as:
\begin{equation}
SE(S_i) = \left( \mu_{L_1}(S_i) - \mu_{L_2}(S_i) \right)^2.   
\end{equation}
\noindent The auxiliary reconstruction loss is defined as:
\begin{equation}
L_{aux-reco} = \sum_{i=1}^{n} SE(S_i)
\end{equation}
\noindent where $S_1$ to $S_n$ represent all the traversable mask labeled in the batch. The main loss is defined as,
\begin{equation}
    L_{main} = \frac{1}{|I|} \sum_{(x,y) \in I} \sum_{j=1}^{4}((I(j,x,y) - O(j,x,y))^2,
\end{equation}
where $L_1$ is the first latent space, and $L_2$ is the second latent space. $I$ is the input data and $O$ is the output data, which are both of size $4 \times H \times W$. $|I|$ represents the total number of pixels in the batch. Finally, the total loss is written as:
\begin{equation}
    L_{total-confidence} = L_{aux-reco} + L_{main}
\end{equation}

\begin{figure}[h]
\centering
\includegraphics[width=0.48\textwidth]{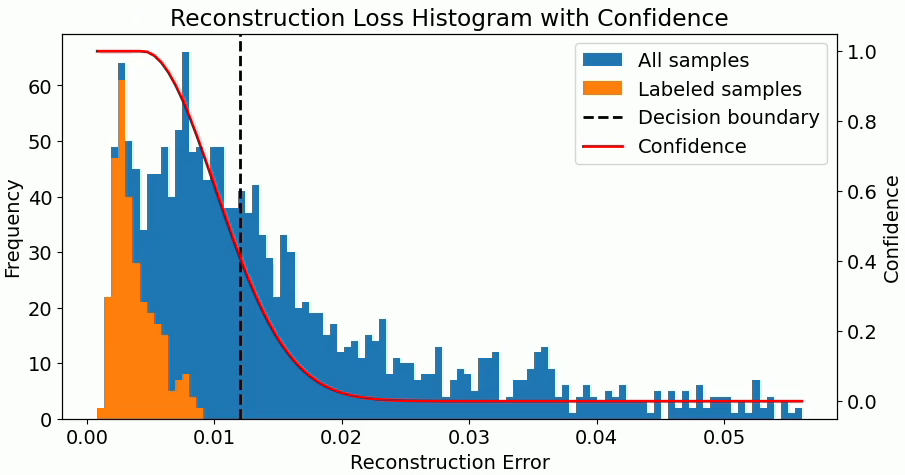}
\caption{Reconstruction histogram of the mask loss $SE(S_i)$ for a batch. Traversable elements are referred to as labeled samples and are shown in orange, while unlabeled elements are depicted in blue. A fixed threshold has been set to label highly reconstructed elements as non-traversable, as explained in \cite{frey2023fast}. The vertical line represents the decision boundary.}
\label{fig:lossDistribution}
\end{figure}

Figure \ref{fig:lossDistribution} shows the distribution of mask loss $SE(S_i)$ for a batch of images after the model have been trained on the dataset.
As expected the labeled samples are correctly reconstructed. For the unlabeled samples, we observe a bimodal distribution, indicating that the traversable parts of the unlabeled samples are being reconstructed well, while the non-traversable parts are not. Thus, with high probability, we can label samples with reconstruction error ($SE(S_i)$) beyond a manually tuned decision boundary as non-traversable. The decision boundary location is selected using a confidence based method similar to  \cite{frey2023fast}. This produces a confidence curve (in red on Fig. \ref{fig:lossDistribution}) that can be used to determine the location of the decision boundary. For more details, the reader is referred to \cite{frey2023fast}. This label
extension method results in another 35.95\% of the image
being assigned labels

\subsection{Model Selection}

The task of COT estimation is closely related to semantic segmentation, which predicts class labels for a predefined set of classes given arbitrary input. 

From scientific literature, three models were selected. All considered models are designed for RGBD inputs to ensure a fair comparison. For each model, we removed the softmax layer typically used for semantic segmentation and directly used a ReLU activation function for regression because COT is an unbounded metric.
A custom baseline model was designed using a CNN Unet architecture \cite{unet}.

The following models were considered:
\begin{itemize}
\item \textbf{Baseline Model}: A CNN Unet architecture. The backbone utilizes ResNet50 to efficiently encode input data into a latent representation.
\item \textbf{DFormer} \cite{dformer}: DFormer consists of a sequence of blocks that encode both RGB and depth information together for enhanced feature extraction. It performs exceptionally well on tasks such as scene segmentation and object detection.
\item \textbf{CMX} \cite{CMX}: CMX (Cross-Modal Fusion) introduces a new framework for RGB-X semantic segmentation. "X" here refers to exploiting features from an arbitrary modality, such as depth.
\item \textbf{AsymFormer} \cite{asymformer}: Designed for understanding indoor scenes, AsymFormer uses RGBD as input data and leverages an asymmetrical backbone for multimodal feature extraction to reduce complexity. Both modalities’ features are fused using LAFS (Local Attention-Guided Feature Selection) and a CMA (Cross-Modal Attention Guided Feature Correlation Embedding) block introduced by the authors.
\end{itemize}

\subsection{Loss Selection}

For our work, we considered a MAE (Mean Absolute Error) loss between the label and the prediction. This choice is justified because the self-supervised process can produce incorrect labels, which could confuse the model during training and MAE loss is less sensitive to outliers compared to MSE loss. 
We use labels after our label generation and label augmentation have been applied.

The MAE loss for a specific pixel is defined as:
\begin{multline}
MAE(b,x,y) =\\
\begin{cases*}
|Z(b,x,y) - P(b,x,y)| & \text{if } $Z(b,x,y) \neq 0$ \\
0 & \text{if  $Z(b,x,y) = 0$ }
\end{cases*}
\end{multline}
where \( B \) is the batch size, \( Z \) corresponds to the labels of size \( B \times H \times W \). An unknown label is represented by $Z(b,x,y) = 0$ and a COT value is represented by a strictly positive number. \( P \) corresponds to the model prediction of size \(
 B \times H \times W \).
 
The total loss for training our RGBD COT deep learning model is defined as follows:
\begin{equation}
L_{total-COT}(Z, P) = \frac{1}{|Z|} \sum_{(b,x,y) \in Z} MAE(b,x,y)
\end{equation}
where $|Z|$ corresponds to the number of pixels in a batch.

\subsection{Heuristic Merger}

Once the pixel-wise COT predictions are generated, these predictions of size \( 1 \times H \times W \) are projected into a local BEV map using the camera's intrinsic and extrinsic parameters. A heuristic global map merger then merges all these local maps, produced sequentially, into a global BEV for path planning.

The global map merger uses the robot’s position to place each local BEV map on a global grid of size \( 2 \times M \times N \). \( M \) and \( N \) are arbitrary sizes representing the size of the map. The first channel corresponds to COT values, and the second channel corresponds to the closest distance at which the cell of the grid has been updated. The merging between the local and global map occurs as follows: each BEV map contains the distance from the camera at the moment the map was generated, and each cell is updated only if the new depth distance is closer than the one already present in the global map cell. The reasoning is that depth estimation and the RGBD-COT model are more accurate when elements such as roads and grass are closer to the camera in the RGBD images.
\section{Results \& Experiments}
In this section, we present the results of our work. A dataset was collected, and a quantitative and qualitative analysis was performed. Finally, an ablation study was conducted to demonstrate the advantage of the proposed model architecture.

\subsection{Data Collection \& Generation}
The M4 robot was driven around the Caltech campus across various terrains, including rocks, roads, and grass. During this time, we collected RGBD frames, position, speed, current, voltage data, and point clouds. The RGBD frames were processed using a label generation method to create COT label images. At this point, only 34.06\% of the dataset was labeled. Following this, we applied the two label augmentation methods described previously, increasing the labeled data to 88.07\%. In total, 743 COT image labels were generated. These labels were split into three parts: 80\% (594) for training, 10\% (74) for validation, and 10\% (75) for testing.
\begin{figure*}[ht]
\centering
\includegraphics[width=\textwidth
]{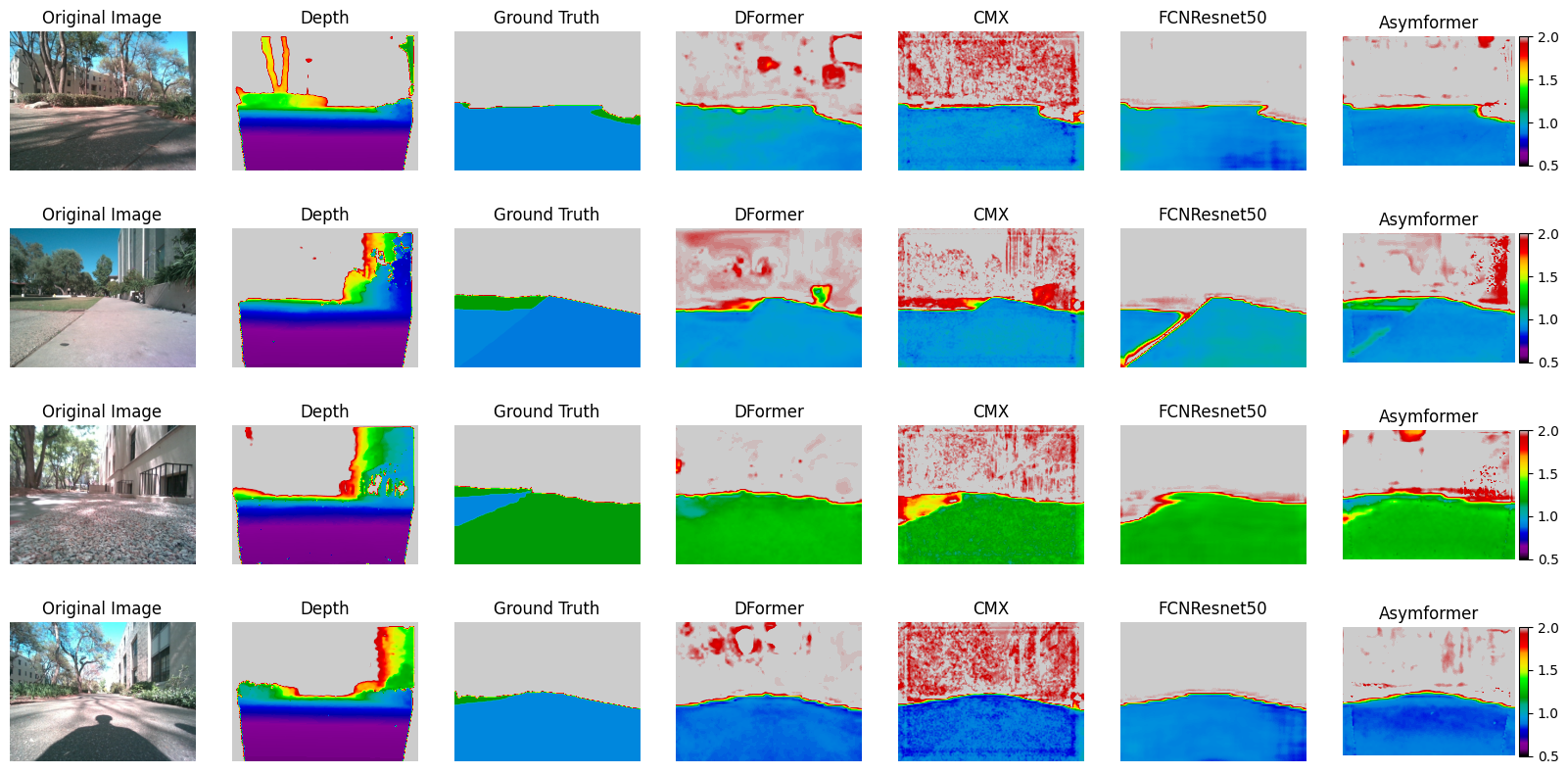}
\caption{Inference on four images of the test dataset. The first two columns are the input given to the model. The third column represents the ground truth manually labeled. The last four columns show the COT images from the models considered. These COT images have been colored from a scale of 0.5 (purple) to 2.0 (gray).}
\label{fig:quali}
\end{figure*}

\subsection{Implementation Details}
Training was performed on a NVIDIA RTX A5000 with 24 GB of memory, running Ubuntu 20.04, CUDA 12.5, and PyTorch 2.3.0 \cite{pytorch}. Data augmentation was applied to the RGBD inputs, including random horizontal flip, color jitter, Gaussian blur, and random resize crop. Due to the nature of the dataset generation using the projected path, only a few images in our dataset contain two terrains labeled side by side. We introduced a copy-paste mask data augmentation method inspired by \cite{copypaste}. In each batch, we randomly selected pairs of RGBD images and their corresponding COT images multiple times. We then pasted a random COT mask from the first image onto the second image. This process created more contact boundaries between different COT terrains, such as grass and road, enhancing the diversity of the training data. For our model backbones, we used pre-trained weights from ImageNet-1k \cite{imagenet} for the ResNet backbone and from the Mix Transformer-B0 \cite{segformre}.

The AdamW optimizer was chosen with a weight decay of 0.01. The learning rate was kept constant at 1e-4. The training loss presented in the method section was used. The autoencoder for confidence generation used two ResNet18 encoders and two custom decoders composed of 2D transposed convolution and Relu activation functions. The channel dimensions were \( c1 = 64 \) and \( c2 = 128 \). The images had a size of 480x640 (HxW) for both confidence generation and model training. For our experiments we used the M4 robot, equipped with a Jetson Orin Nano and a D455 realsense camera. The inference pipeline is implemented in Python using TensorRT and ROS1 Noetic \cite{ros}. After optimizing of the model using TensorRT, the full pipeline presented in Figure \ref{fig:overallpipeline} was able to run at 4fps on the Jetson Orin Nano.

\subsection{Qualitative Analysis}
After training all the models, we conducted a qualitative analysis. We manually labeled the test dataset by assigning each terrain, the average COT for that specific terrain computed over the whole dataset. Figure \ref{fig:quali} displays the performance of the models on four selected images from the test dataset. All models succeeded in recognizing the appropriate COT for each terrain.
However, the baseline, DFormer, and CMX models had difficulties distinguishing clear boundaries, as shown in the second row. Only the Asymformer model managed to do so, as shown in the second row. Overall, from a qualitative point of view, the Asymformer model seems to be the best candidate for our task.

\subsection{Quantitative Analysis}

To make the model comparison quantitative, we compared the performance of the models using the MSE (Mean Squared Error) loss. The results are summarized in Table \ref{tab:comp}. We chose MSE as the evaluation metric in order to impose significant penalties on models that inaccurately assign distant COT values. For instance, this metric heavily penalizes models that incorrectly classify a traversable terrain as non-traversable.

\begin{table}[h!]
\centering
\renewcommand{\arraystretch}{1.2}
\begin{tabular}{l|cccc}
\hline
\textbf{Metric} &  \textbf{Baseline} & \textbf{DFormer} & \textbf{CMX} & \textbf{Asymformer} \\ \hline
MSE (unitless) & 0.0333 & 0.0333 & 0.0291 &\textbf{ 0.0252} \\ 
Inference time (ms)&\textbf{ 16.1 }& 45.1 & 76.2 & 19.8 \\ \hline

\end{tabular}
\caption{First row: performance of the models on the test dataset using the
MSE (Mean Squared Error) metric pixel-wise (lower is better).
Second row: inference time of the models on a NVIDIA RTX A5000 using Pytorch \cite{pytorch} (lower is better).
The bold value represents the best model for the metric.}
\label{tab:comp}
\end{table}

Our method used is referenced as C-SAM standing for Confidence - Segment Anything Model. We utilize both label augmentation methods explained in the method section. Again, the Asymformer model appears to be the most promising candidate, achieving the lowest MSE score and the smallest standard deviation. Other models achieved performance close to Asymformer, deviating by approximately 30\% in MSE metric.

A significant difference between these models lies in their inference time, as illustrated in Table \ref{tab:comp}. Asymformer and the Baseline demonstrates greater efficiency compared to the other models examined, being four times more efficient than the CMX model. For our practical deployment and real-world testing, we selected Asymformer as the appropriate model for our task.

\subsection{Method Ablation}

To prove that the proposed method achieves excellent performance in generating reliable training labels, other labeling methods were considered. All models were retrained for each case to assess their reliability.

\begin{itemize}
\item \textbf{SL}: Standard Loss, using only the labels from the projected path without SAM and Confidence extension. This is our baseline case.
\item \textbf{SL-SAM}: Standard Loss SAM, extending from the previous case by adding the labels from SAM.
\item \textbf{UN-SAM}: Unlabeled Negative SAM, extending again from SL-SAM by considering all the unlabeled data as non-traversable.
\item \textbf{C-SAM (ours)}: The method explained in this work, using the SAM-extended labels and the confidence-extended labels.
\end{itemize}

\begin{table}[h!]
\centering
\renewcommand{\arraystretch}{1.2}
\setlength{\tabcolsep}{2pt} 
\begin{tabular}{l|cccc|c}
\hline
\textbf{Method} & \textbf{Baseline} & \textbf{DFormer} & \textbf{CMX} & \textbf{Asymformer} &\textbf{ \% Labeled} \\ \hline
SL & 0.0444 & 0.1278 & 0.0573 & \textbf{0.0437} & 34.06 \\
SL-SAM & 0.0408 & 0.1288 &\textbf{ 0.0388} & 0.0405 & 52.12\\
UN-SAM & 0.0397 & 0.0371 & 0.0423 & \textbf{0.0394} & 100.00\\
C-SAM (ours)  & 0.0333 & 0.0333 & 0.0291 &\textbf{ 0.0252} & 88.07  \\\hline
\end{tabular}
\caption{Comparison of MSE metric on the test dataset for the different cases. The bold value represents the best model for this method. The "\% Labeled" column represents the percentage of labeled data in the dataset.}
\label{tab:mean_values}
\end{table}
From Table \ref{tab:mean_values}, our C-SAM method outperforms the previous cases considered, improving model performance by up to 74\% for the DFormer and of 42.2\% the Asymformer model. The SL case contains only a small amount of labeled data per training image, resulting in poor performance. SL-SAM performs outperforms SL, by adding the labels from SAM. UN-SAM contains fully labeled images; however, some labels are incorrectly annotated, which confuses the model due to the assumption that unlabeled data is non-traversable. Lastly, the confidence label augmentation method allows the C-SAM method to have a more accurate labeling of the unlabeled data than UN-SAM. This allows the method to outperforms all others despite having 12\% less labels compared to the UN-SAM method.

\subsection{Practical Deployment and Real-World Testing}
To demonstrate the usefulness of our method, an A* Algorithm was implemented and executed on the map created using our approach, as shown in Figure \ref{fig:results}. Two paths are displayed for traveling from point A to point B. Assuming this map as prior knowledge, the red path represents the most optimal route, minimizing the COT over the path. The yellow path on the right, despite being shorter in distance than the red one, has a higher accumulated COT. The results demonstrate that generating a COT map for robotic navigation is critical for efficient movement.

\section{Discussion \& Conclusion}
In this work we have designed, implemented, and demonstrated a novel method for traversability estimation based on the cost of transport metric. To achieve this, we incorporated light assumptions and two data augmentation methods to label raw data, and create a suitable dataset for our RGBD COT model. This model predicts 2D COT images and uses the camera's intrinsic and extrinsic parameters to project this information into 3D space. These projections are then fused into a global BEV map using our heuristic map merger. 

We demonstrated the performance of our model both in simulation and in real-world field tests. During data collection, the commanded speed, meaning the desired speed set for the controller, was kept constant. It was observed that the energy consumed by the robot primarily increases when the robot wheels slip. However, when the commanded speed varies, energy consumption can be affected by other factors such as motor inefficiency. Therefore, developing a method to handle varying commanded speeds is an area for future research.

Further improvements could involve replacing the heuristic map merger with a 2D temporal deep learning network, potentially enhancing the integration of the COT map into a global map. The success of the entire work heavily relies on the quality of the SAM mask segmentation. Thus, exploring alternative models for image segmentation could also be a promising direction for improvement.

Lastly, we could leverage the flying capabilities of the M4 robot to generate a map of the safest areas for landing using self-supervised methods. Additionally, taking advantage of the various ground modes of locomotion of the M4 based on the COT global map could enable the robot to autonomously adapt to different environments. For instance, the robot could crawl on rocks or sand, or drive normally on pavements. Such adaptability would significantly enhance the M4 robot's autonomy, providing a substantial advantage in diverse terrains.









\addtolength{\textheight}{-12cm}   





\section*{ACKNOWLEDGMENT}

We would like to acknowledge funding from the Center for Autonomous Systems and Technology (CAST) at Caltech.

\end{document}